\documentclass[10pt,twocolumn,letterpaper]{article}

\usepackage{iccv}
\usepackage{times}
\usepackage{epsfig}
\usepackage{graphicx}
\usepackage{amsmath}
\usepackage{amssymb}

\usepackage{arydshln}
\usepackage{booktabs}
\usepackage{graphicx}
\usepackage{multirow}
\usepackage{tabu}
\usepackage{siunitx}
\usepackage[dvipsnames,svgnames,x11names]{xcolor}
\usepackage{subfig}
\usepackage{adjustbox}
\usepackage[demo,abs]{overpic}
\usepackage{enumitem}
\setlist[itemize]{align=parleft,left=0pt..1em}

\usepackage{soul}
\usepackage{capt-of}

\usepackage[pagebackref=true,breaklinks=true,letterpaper=true,colorlinks,bookmarks=false]{hyperref}
\hypersetup{
  linkcolor = darkred,  
  citecolor = darkgreen 
}

\definecolor{darkgreen}{RGB}{30,150,30}
\definecolor{darkblue}{RGB}{0,0,127}
\definecolor{darkyellow}{RGB}{171,133,0}
\definecolor{darkred}{RGB}{180,20,20}
\definecolor{darkmagenta}{RGB}{127,0,127}
\definecolor{darkcyan}{RGB}{0,127,127}
\definecolor{orange}{rgb}{0.8, 0.4, 0.0}

\iccvfinalcopy

\newif\ifdrafting 
\draftingtrue 

\newcommand{\comment} [1] {}

\def\iccvPaperID{1636} 

\ificcvfinal\fi

\begin{document}

\title{SLIDE: Single Image 3D Photography with \\ Soft Layering and Depth-aware Inpainting}

\author{Varun Jampani$^*$, Huiwen Chang$^*$, Kyle Sargent, Abhishek Kar, Richard Tucker, Michael Krainin, \\ 
Dominik Kaeser, William T. Freeman, David Salesin, Brian Curless, Ce Liu\\
\vspace{4ex}
{Google} \\
\vspace{-4ex}
}

\twocolumn[{%
\vspace{-5mm}
\renewcommand\twocolumn[1][]{#1}%
\maketitle

\begin{center}
    \newcommand{\teaserwidth}{\textwidth}
    \vspace{-4mm}
    \centerline{
    \includegraphics[width=\teaserwidth]{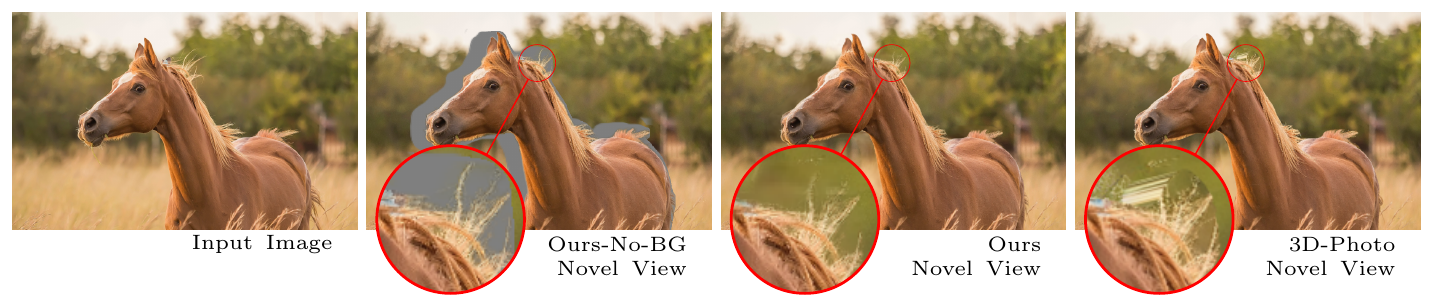}
    }
    \captionof{figure}{\textbf{Appearance Details with SLIDE.} 
    View synthesis results show better preservation of hair structures in SLIDE (Ours) compared to that of 3D-Photo~\cite{shih20203d}.
    We also show the novel view (Ours-No-BG) where the background (BG) layer is greyed-out
    to showcase our soft layering. See the supplementary video for a better illustration
    of view synthesis results.}
    \vspace{-2mm}
	\label{fig:teaser}
\end{center}%
}]

\footnotetext[1]{Equal Contribution.}


\begin{abstract}
\vspace{-2mm}
Single image 3D photography enables viewers to view a
still image from novel viewpoints.
Recent approaches combine
monocular depth networks with inpainting networks
to achieve compelling results.
A drawback of these techniques is the use of hard depth layering,
making them unable to model intricate appearance
details such as thin hair-like structures.
We present SLIDE, a modular and unified system for single
image 3D photography that uses a simple yet effective soft
layering strategy to better preserve appearance details in novel views.
In addition, we propose a novel depth-aware training strategy for
our inpainting module, better suited for the 3D photography task.
The resulting SLIDE approach is modular, enabling the use
of other components such as segmentation and matting for improved layering.
At the same time, SLIDE uses an efficient layered depth formulation 
that only requires a single forward pass through the component networks to produce high quality 3D photos.
Extensive experimental analysis on three view-synthesis
datasets, in combination with user studies on in-the-wild image collections,
demonstrate superior performance of our technique in comparison
to existing strong baselines while being conceptually much simpler.
Project page: \url{{https://varunjampani.github.io/slide}}
\vspace{-3mm}
\end{abstract}


\section{Introduction}
\label{sec:intro}
\vspace{-2mm}

Still images remain a popular choice for capturing, storing, and sharing 
visual memories despite the advances in richer capturing technologies such
as depth and video sensing. Recent advances~\cite{tucker2020single,wiles2020synsin,niklaus:2019:kenburns,shih20203d,kopf2019practical,kopf2020one} 
show how such 2D images can be ``brought to life'' just by interactively changing the camera viewpoint, even without scene movement, thereby creating a more engaging 3D viewing experience.
Following recent works, we use the term `Single image 3D photography' to describe the process of converting a 2D image into a 3D viewing experience.
Single image 3D photography is quite challenging, as it requires estimating scene geometry from a single image along with inferring the dissoccluded scene
content when moving the camera around.
Recent state-of-the-art techniques for this problem can be broadly classified into two approaches - \textit{modular systems}~\cite{niklaus:2019:kenburns, shih20203d} 
and \textit{monolithic networks}~\cite{tucker2020single, wiles2020synsin}. 

Modular systems~\cite{shih20203d,niklaus:2019:kenburns,kopf2019practical,kopf2020one} 
leverage state-of-the-art
2D networks such as single-image depth estimation, 2D inpainting, and instance segmentation.
Given recent advances in monocular depth estimation~\cite{lasinger:2019:depth_3d,li:2018:megadepth, li:2019:moving_people} and inpainting~\cite{yi:2020:contextual_hifill,yu:2018:generative_inpainting}
fueled by deep learning on large-scale 2D datasets,
these modular approaches have been shown to work remarkably well on in-the-wild images.
A key component of these modular approaches is decomposing the scene into a set of
layers based on depth discontinuities. The scene is usually decomposed
into a set of layers with \emph{hard} discontinuities and thus can not model soft appearance effects
such as matting. See Figure~\ref{fig:teaser} (right) for an example novel view
synthesis result from 3D-Photo~\cite{shih20203d}, which is a state-of-the-art
single image 3D photography system.

In contrast, monolithic approaches~\cite{tucker2020single, wiles2020synsin}
attempt to learn end-to-end trainable networks using view synthesis
losses on multi-view image datasets. These networks usually take a single image
as input and produce a 3D representation of a scene, such as point clouds~\cite{wiles2020synsin}
or multi-plane images~\cite{tucker2020single}, from which one could interactively render the scene from
different camera viewpoints. Since these networks usually decompose the scene
into a set of soft 3D layers~\cite{tucker2020single} or directly generate 3D structures~\cite{wiles2020synsin},
they can model appearance effects such as matting. Despite being elegant, these networks usually perform poorly
while inferring disoccluded content and have difficulty generalizing
to scenes out of the training distribution, a considerable limitation given the difficulty in obtaining multi-view datasets on a wide range of scene types.

In this work, we propose a new 3D photography approach that uses soft layering and
depth-aware inpainting. We refer to our approach as `SLIDE' (\ul{S}oft-\ul{L}ayering
and \ul{I}npainting that is \ul{De}pth-aware). Our key technique is a simple
yet effective soft layering scheme that can incorporate intricate
appearance effects.
See Figure~\ref{fig:teaser} for an example view synthesis result of SLIDE (Ours), where
thin hair structures are preserved in novel views.
In addition, we propose an RGBD inpainting network
that is trained in a novel depth-aware fashion resulting in higher quality view synthesis.
The resulting SLIDE framework is modular, and allows easy incorporation
of state-of-the-art components such as depth and segmentation networks.
SLIDE uses a simple two-layer decomposition of the scene and requires only a single
forward pass through different components. This is in
contrast to the state-of-the-art 
approaches~\cite{shih20203d,niklaus:2019:kenburns}, which are modular and 
require several passes through some components networks.
Moreover, all of the components in the SLIDE framework are differentiable and can be 
implemented using standard GPU layers in a deep learning toolbox, resulting in a unified system. 
This also brings our SLIDE framework closer to single network approaches.
We make the following contributions in this work:
\begin{itemize}[leftmargin=*]
    \vspace{-2mm}
    \item We propose a simple yet effective soft layering formulation that enables synthesizing intricate appearance details such as thin hair-like structures in novel views.
    \vspace{-2mm}
    \item We propose a novel depth-aware technique for training an inpainting network for the 3D photography task.
    \vspace{-2mm}
    \item The resulting SLIDE framework is both modular and unified with favorable properties such as only requiring a single forward computation with favorable runtime.
    \vspace{-2mm}
    \item Extensive experiments on four different datasets demonstrate the superior performance of SLIDE in terms of both quantitative metrics as well as from user studies.
\end{itemize}

\section{Related Work}
\label{sec:related}
\vspace{-2mm}

Classical optimization methods have been applied to the view synthesis task \cite{gortler:1996:lumigraph, levoy:1996:lightfield, penner:2017:soft3d}, but the most recent approaches are learning-based. Some works~\cite{flynn:2016:deepstereo, kalantari:2016:learning} have predicted novel views independently, but to achieve consistency between output views, it is preferable to predict a scene representation from which many output views can be generated. Such representations include point clouds~\cite{wiles2020synsin, meshry:2019:neural_rerendering}, meshes~\cite{shih20203d}, layered representations such as layered depth images~\cite{shade:1998:ldi,tulsiani:2018:lsi} and multi-plane images (MPIs)~\cite{zhou2018stereo,flynn:2019:deepview,srinivasan:2019:boundaries}, and implicit representations such as NeRF~\cite{mildenhall:2020:nerf, martinbrualla:2020:nerf_wild,boss2021nerd}.
Much research in view synthesis has centered on the task of interpolation between multiple images, but most relevant to our work are methods focusing on the very challenging task of extrapolation from a single image.

\vspace{1mm}
\noindent \textbf{Single Network Approaches.}
For narrow baselines, Srinivasan \etal\ predict a 4D lightfield directly \cite{srinivasan:2017:lightfield}, while Li and Kalantari~\cite{li:2020:lightfield_variable_mpi_vmpi} represent a lightfield as a blend of two Variable-depth MPIs. For larger baselines, Single-view MPI~\cite{tucker2020single} applies the MPI representation to the single image case, and SynSin~\cite{wiles2020synsin}
uses point clouds and applies a neural rendering stage, which enables it to generate content outside the original camera frustum.
These learning-based methods are trained end-to-end, with held-out views from novel viewpoints being used for supervision via a reconstruction loss. Training data can be obtained from light-field cameras~\cite{li:2020:lightfield_variable_mpi_vmpi,srinivasan:2017:lightfield} or multi-camera rigs~\cite{flynn:2019:deepview}, or derived from photo collections~\cite{meshry:2019:neural_rerendering} or videos of static scenes~\cite{zhou2018stereo}. 
A key drawback of these approaches is their poor generalization to in-the-wild images.

\begin{figure*}[t!]
\centering
\includegraphics[width=.95\textwidth]{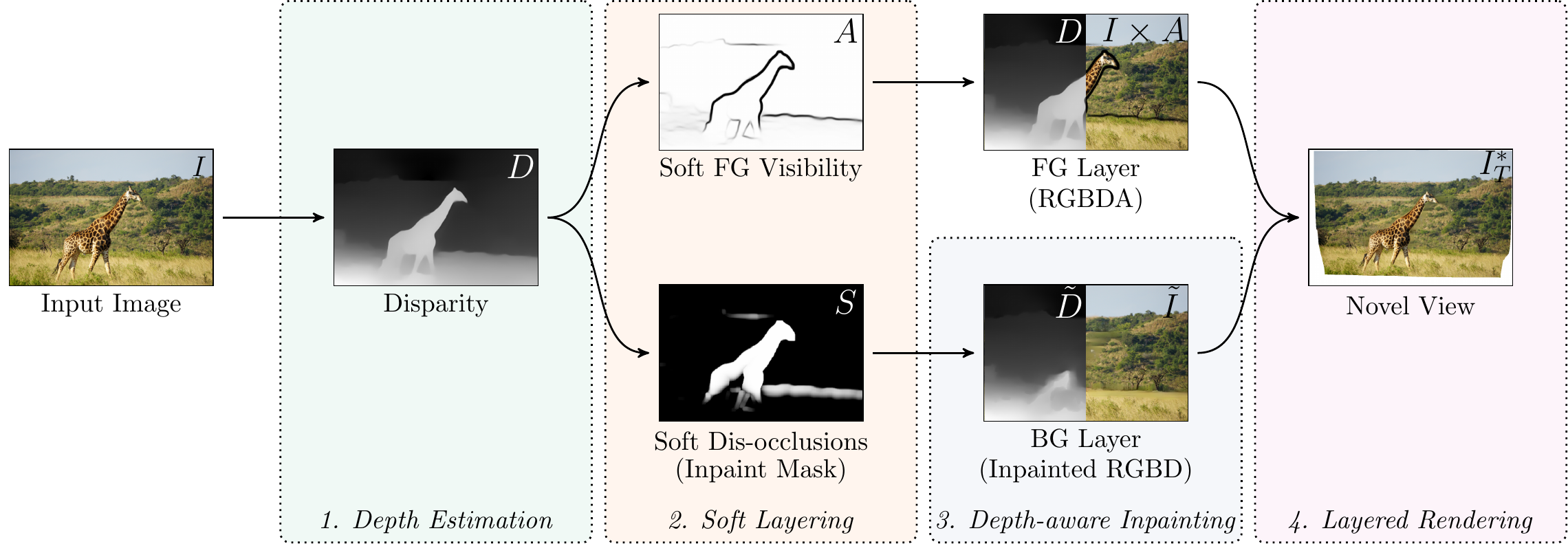}
\vspace{.5mm}
\caption{\textbf{SLIDE Overview.} SLIDE is a modular and unified framework for 3D photography
and consists of the four main components of depth estimation, soft layering, depth-aware RGBD inpainting and layered rendering. In addition, one can optionally use foreground alpha mattes (not shown in this figure) to improve the layering.}
\vspace{-1.5mm}
\label{fig:pipeline}
\end{figure*}

\vspace{1mm}
\noindent \textbf{Depth-based 3D Photography.}
Another approach to single image 3D photography is to build a system combining depth prediction and inpainting modules 
with a 3D renderer.
For depth estimation from a single image, a variety of learning-based methods~\cite{chen:2016:depth_in_the_wild_diw,li:2018:megadepth,garg:2016:unsupervised} exist.
The MiDaS system~\cite{lasinger:2019:depth_3d} achieves excellent results by training on frames from 3D movies~\cite{lasinger:2019:depth_3d} as well as other depth datasets.
Non-learning-based approaches for inpainting apply patch matching and blending \cite{barnes:2009:patch_match,huang:2014:image_completion} or diffusion \cite{bertalmio:2000:image_inpainting,weickert:1999:coherence_diffusion}, but since arbitrary amounts of training data can be generated simply by obscuring random sections of input images, this task is a natural target for learning-based approaches. Recent methods propose augmenting convolutional networks with contextual attention mechanisms such as DeepFill's gated convolutions \cite{yu:2018:generative_inpainting,yu:2019:gated_convolution_inpainting} or a coherent semantic attention layer \cite{liu:2019:coherent_inpainting}, and applying patch-based discriminators for GAN-based training as well as reconstruction losses. Another recent work, HiFill~\cite{yi:2020:contextual_hifill} takes a residual-based approach to inpaint even very high resolution images.

In the context of 3D Photography, inpainting will typically operate on a more complex representation than a simple image, and systems may need to inpaint depth as well as texture. The method of Shih \etal\ ~\cite{shih20203d} introduces an extension of the layered depth image format. The system performs multiple inpainting steps in which edges and depth, as well as images, are inpainted within different image patches. The system of Niklaus \etal\ ~\cite{niklaus:2019:kenburns} performs inpainting on rendered novel images and projects the inpainted content back into a point cloud to augment its representation. This latter system also adds an additional network to refine the estimated depth, and incorporates instance segmentation to ensure that people and other important objects in the scene do not straddle depth boundaries. 
These systems are somewhat complex, requiring multiple passes of the same network (e.g., inpainting).
As mentioned earlier, another key drawback of these approaches is that the layering is hard and can not incorporate intricate appearance effects in layers.
Our approach follows the depth-plus-inpainting paradigm but operates on a simple yet effective two-layer representation that enables the incorporation of soft appearance effects. In addition, due to our simple two-layer soft formulation, we only need
a single forward pass through different component networks; thus, it can be considered as a single unified network while being modular.

\section{Methods}
\label{sec:methods}
\vspace{-2mm}

\vspace{1mm}
\noindent \textbf{SLIDE Overview.}
As illustrated in Figure~\ref{fig:pipeline}, our 3D photography approach,
SLIDE, has four main components: 1.~monocular depth estimation,
2.~soft layering, 3.~depth-aware RGBD inpainting,
and 4.~layered rendering.
From a given still image $I \in \mathbb{R}^{n \times 3}$ with $n$ pixels, 
we first estimate depth $D \in \mathbb{R}^n$. We then decompose the scene
into two layers via our soft-layering formulation where we estimate
foreground (FG) pixel visibility $A \in \mathbb{R}^n$ and inpainting 
mask $S \in \mathbb{R}^n$ in a soft manner.
Using these, we construct the foreground RGBDA layer with 
the input image $I$, its corresponding disparity $D$ and the pixel visibility map $A$;
and the background RGBD layer with the inpainted RGB image $\tilde{I}$ and the inpainted 
disparity map $\tilde{D}$.
We then construct triangle meshes from the two disparity maps, textured with $I$ and $A$ for the foreground and $\tilde{I}$ for the background, render each into a target viewpoint, and composite the foreground rendering over the background rendering.
In addition, we can optionally use foreground alpha mattes to improve 
layering as our soft layering enables easy incorporation of alpha-mattes. 

\subsection{Monocular Depth Estimation}
\label{sec:depth}
\vspace{-2mm}
Given an RGB image $I \in \mathbb{R}^{n \times 3}$ with $n$ pixels,
we first estimate a disparity (inverse depth) map $D \in \mathbb{R}^{n \times 1}$ 
using a CNN. We use the publicly released MiDaS v2~\cite{lasinger:2019:depth_3d} network $\mathbf{\Phi}_D$ for monocular inverse depth prediction. Specifically, the MiDaS model 
is trained on a large and diverse set of datasets to achieve zero-shot cross dataset transfer.
It proposes a principled dataset mixing strategy and a robust scale and shift invariant loss 
function that results in predicted disparity maps up to an unknown scale and shift factor.
 The final output of $\mathbf{\Phi}_D$
is a normalized disparity map $D \in [0, 1]^{n}$, which is then used in the subsequent parts
of the SLIDE pipeline. 
To reduce missing foreground pixels and noise in layering (Section~\ref{sec:layering}),
we do slight Gaussian blur and max-pool the disparity map.
One could use any other monocular depth network
in our framework. We choose MiDaS for its good generalization across different types of images.

\subsection{Soft Layering}
\label{sec:layering}
\vspace{-2mm}

A key technical contribution in SLIDE is estimating layers in a soft fashion so that
we can model partial visibility effects across layers. 
As illustrated in Figure~\ref{fig:pipeline}, layering also connects the depth and
inpainting networks making it a crucial component of SLIDE.
Our soft layering has two main components: 1. estimating
soft pixel visibility of the foreground layer, and 2. estimating a soft disocclusion
map that is used for background RGBD inpainting.

\begin{figure}[t!]
\begin{minipage}{\columnwidth}
    \centering
    \includegraphics[width=0.95\columnwidth]{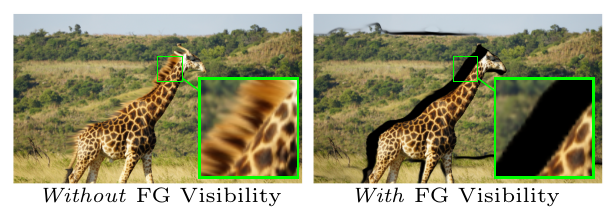}
    \caption{\textbf{Foreground Pixel Visibility.} Rendering an RGBD layer without pixel visibility (left) leads to stretchy triangles, whereas rendering with pixel visibilities (right) enables seeing through to the background (represented with black pixels here).}
    \label{fig:visibility}
    \vspace{-2mm}
\end{minipage}
\end{figure}

\vspace{1mm}
\noindent \textbf{Soft FG Pixel Visibility.}
We estimate visibility at each image pixel, which enables us
to see through to the background layer when rendering novel-view images.
Figure~\ref{fig:visibility} (left) shows a single RGBD layer, represented by the given RGB image and the corresponding
estimated disparity, rendered as a textured triangle mesh into a new viewpoint. Stretching artifacts appear at depth discontinuities.  To address these artifacts, we construct a visibility map $A$ that has lower values (higher transparency) at depth discontinuities -- lower visibility in proportion to changes in disparity -- later allowing us to see through these discontinuities to the (inpainted) background layer.  
More formally, given the estimated disparity map $D$
for a given image $I$, we compute the FG pixel visibility map $A \in [0,1]^n$ as:
\vspace{-2mm}
\begin{equation}
    A = e^{-\beta ||\nabla D||^2},
    \label{eqn:visibility}
    \vspace{-2mm}
\end{equation}
where $\nabla$ is the Sobel gradient operator and $\beta \in \mathbb{R}$ is a scalar parameter.
Thus the pixel visibility varies inversely with disparity gradient magnitude.
Low FG visibility ($A \approx 0$) corresponds to high FG pixel transparency. 
Figure~\ref{fig:visibility} (right) shows a novel-view rendering with the pixel
visibility map $A$ multiplied against the original rendering; black regions indicate areas that are now transparent in the foreground layer.
Note that modelling this FG visibility in a soft manner allows SLIDE to easily incorporate segmentation
based soft alpha mattes into layering, as we discuss
in Section~\ref{sec:segmentation}.

\begin{figure}[t!]
\begin{minipage}{\columnwidth}
    \centering
    \includegraphics[width=\columnwidth]{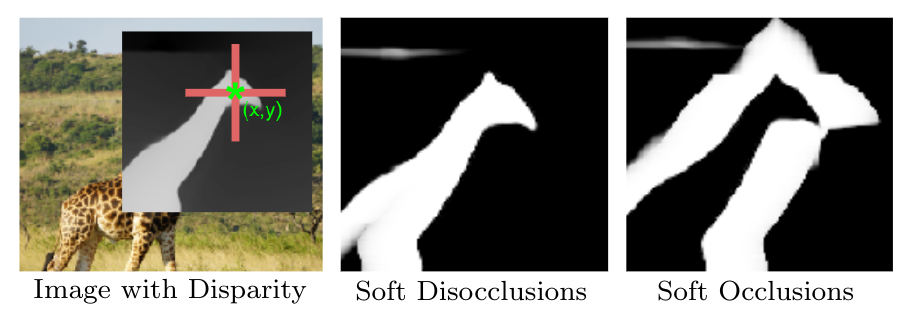}
    \caption{\textbf{Soft Disocclusions and Occlusions.} At each point $(x,y)$ in an image, we compare disparity differences across horizontal and vertical scan lines to compute soft dissocclusion and occlusion maps.}
    \vspace{-2mm}
    \label{fig:occlusions}
\end{minipage}
\end{figure}

\begin{figure*}[t!]
\centering
\vspace{-4mm}
\includegraphics[width=0.9\textwidth]{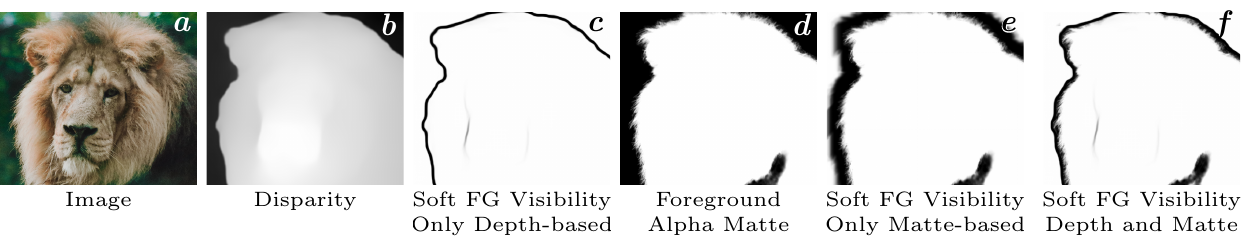}
\caption{\textbf{Layering With Alpha Mattes.} Depth-based FG visibility (c) can not capture hair-like structures. Computing FG visibility (e) based on FG alpha matte (d) and then incorporating it into visibility can capture fine details (f).}
\label{fig:seg_layering}
\vspace{-4mm}
\end{figure*}

\vspace{1mm}
\noindent \textbf{Soft Disocclusions.}
In addition to foreground visiblity, we need to construct a mask to guide inpainting in the background layer. Intuitively, we need to inpaint the pixels
that have potential to be dissoccluded when the camera moves around. 
The relationship between (dis-)occlusion and disparity is well known~\cite{belhumeur1996bayesian,wang2019local,wang2020improving}, and we make use of this relationship to compute soft disocclusions from the estimated disparities.
The disparity-occlusion relationship in these prior works is derived in the stereo 
image setting, where we have metric disparity, and the occlusions are defined with respect
to a second camera. In our case, we only have relative depth (disparity), but we 
can still assume some maximum camera movement and introduce a scalar parameter that can 
scale the disparities accordingly.
Consider the background region at pixel location $(x,y)$ behind the giraffe's
head in Figure~\ref{fig:occlusions}~(left). 
This background region has the potential to be dissoccluded by the foreground if there exists a
neighborhood pixel $(x_i, y_j)$ whose disparity difference with respect to
the foreground pixel at $(x, y)$ is greater than the distance between those pixels' locations.
More formally, the background pixel at $(x,y)$ will be dissoccluded if
\vspace{-1mm}
\begin{equation}
    \exists_{(x_i,y_j)} \left(D_{(x,y)} - D_{(x_i,y_j)} > \rho K_{(x_i,y_j)} \right),
    \label{eqn:disocclusion}
    \vspace{-.7mm}
\end{equation}
where $\rho$ is a scalar parameter that scales the disparity difference. 
$K_{(x_i,y_j)} = \sqrt{(x_i-x)^2 + (y_j-y)^2}$ is
the distance between $(x_i,y_j)$ to the center pixel location $(x, y)$.
In simpler terms, a background pixel is more likely to be dissoccluded 
if the foreground disparity at that point
is higher compared to that of surrounding regions.
For our soft-layering formulation, we convert the above step function into a 
soft version resulting in a soft disocclusion map $S \in [0,1]^n$:
\vspace{-2mm}
\begin{eqnarray}
    S_{(x,y)} = \mathrm{tanh} \Big(\gamma \mathrm{max}_{(x_i,y_j)}\big(&\!\!\!\!\! D(x,y) - D(x_i,y_j) \nonumber\vspace{-.5mm}\\
    &\!\!\!\!\!\!\! - \rho K_{(x_i,y_j)} \big)\Big),
    \label{eqn:soft-disocclusion}
    \vspace{-2mm}
\end{eqnarray}
where $\gamma$ is another scalar parameter that controls the steepness of the tanh activation.
In addition, we apply a ReLU activation on top of tanh to make $S$ positive.
Computing $S$ with the above equation requires computing pairwise disparity differences
between all the pixel pairs in an image. 
Since this is computationally expensive, we constrain the
disparity difference computation to a fixed neighborhood ($m$ pixels) around each
pixel ($(x_i,y_j) \in \mathcal{N}_{(x,y)}$, 
where $\mathcal{N}$ is the $m$ pixel neighborhood of $(x,y)$).
This is still computationally expensive for reasonable values of $m$ ($>30$).
So, we constrain our disparity difference computation along horizontal and 
vertical scan lines as illustrated with red lines in Figure~\ref{fig:occlusions}~(left).
We implement pairwise disparity differences and also pixel distances along
horizontal and vertical neighborhoods with efficient convolution operations
on disparity and pixel coordinate maps.
This results in an effective feed-forward computation of dissocclusion maps using standard
deep learning layers.
For efficiency, we can also compute the disocclusion map 
on a downsampled disparity map and then upsample the resulting map to the desired
resolution.
Figure~\ref{fig:occlusions}~(middle) shows the soft dissocclusion map that is estimated
using this technique.
In a similar fashion, we obtain a soft {\em occlusion} map $\hat{S} \in [0,1]^n$ as shown in
Figure~\ref{fig:occlusions}~(right) by replacing `$>$' with `$<$' in Eqn.~\ref{eqn:disocclusion}
and modifying Eqn.~\ref{eqn:soft-disocclusion} accordingly.
We make use of the occlusion and disocclusion maps in inpainting training (Section~\ref{sec:inpainting}).

\subsection{Improved Layering with Segmentation}
\label{sec:segmentation}
\vspace{-2mm}

Consider the input image shown in Figure~\ref{fig:seg_layering}(a), which has thin hair structures.
The soft FG visibilities (Eqn.~\ref{eqn:visibility}) purely based on depth 
discontinuities will not preserve such thin structures in novel views. For this image, we can see that these fine structures are not captured in the disparity map and are therefore also missed in the visibility map (Figure~\ref{fig:seg_layering}~(b) and (c), resp.).
To address this shortcoming, we incorporate FG alpha mattes -- computed using state-of-the-art
segmentation and matting techniques -- into our layering.
Our soft layering can naturally incorporate soft mattes.

We first compute FG segmentation using the U2Net saliency network~\cite{qin2020u2}, which we then
pass to a matting network~\cite{forte2020f} to obtain FG alpha matte $M \in [0,1]^n$ as shown in Figure~\ref{fig:seg_layering}~(d).
Note that we can not directly use these alpha mattes as visibility maps, as we want visibility
to be low (close to zero) \textit{only} around the FG object boundaries.
So, we dilate (max-pool) the alpha matte (denoted as $\bar{M}$) and then subtract the original alpha matte from it. 
Figure~\ref{fig:seg_layering}~(e) shows $1-(\bar{M}-M)$.
The resulting matte-based FG visibility map will only have low visibility around FG alpha matte.
We then compute the depth-matte
based FG visibility map $A' \in [0,1]^n$ as:
$A' = A * (1 - (\bar{M}-M)(1-\hat{S}))$,
where $A$ denotes the depth-based visibility map (Eqn.~\ref{eqn:visibility}, Figure~\ref{fig:seg_layering} (c))
and $\hat{S}$ denotes the occlusion map, with example shown in 
Figure~\ref{fig:occlusions} (right).
$(1-\hat{S})$ term reduces the leakage of matte-based visibility map onto too
much background and the multiplication with depth-based visibility $A$ ensures
that the final visibility map also accounts for depth discontinuities.
Figure~\ref{fig:seg_layering} (f) shows the depth-matte based FG visibility map $A'$,
which captures fine hair-like structures while also respecting the depth discontinuities. 

\begin{figure*}[t!]
\centering
\vspace{-4mm}
\includegraphics[width=.9\textwidth]{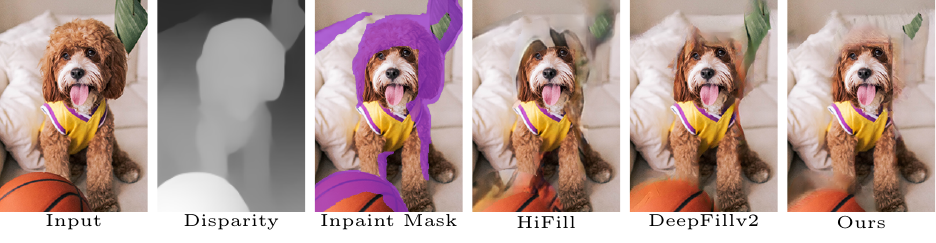}
\caption{\textbf{Depth-aware Inpainting.} Inpainting techniques (HiFill~\cite{yi:2020:contextual_hifill} and DeepFill~\cite{yu:2019:gated_convolution_inpainting}) borrow information from both FG and BG. Our depth-aware inpainting mostly borrows information from the BG making it more suitable for 3D photography.}
\label{fig:depth-aware-inpainting}
\vspace{-4mm}
\end{figure*}

\subsection{Depth-Aware RGBD Inpainting}
\label{sec:inpainting}
\vspace{-2mm}
To avoid exposing black background pixels when the camera moves, as shown in Figure~\ref{fig:visibility},  we inpaint the disoccluded regions $S$ and incorporate the result into our background layer.  Inpainting such disocclusions is different from traditional inpainting problems because the model needs to learn to neglect the regions in front of each to-be-inpainted pixel. In Figure~\ref{fig:depth-aware-inpainting}, we show sample results of two state-of-the-art image inpainting methods~\cite{yu:2019:gated_convolution_inpainting, yi:2020:contextual_hifill} to inpaint the disoccluded regions.
While they synthesize good textures and are even capable of completing the basketball and the dog's head, this is actually undesirable in our pipeline, as we want to inpaint the BG, not the FG. In addition, we perform RGBD inpainting which is in contrast to the existing RGB inpainting networks.

One of the key challenges in training our inpainting network using disocclusion masks (see Figure~\ref{fig:occlusions} (middle) for an example mask) is that we do not have ground-truth background RGB or depth for these regions in single image datasets. To overcome this, we instead make use of occlusion masks (see Figure~\ref{fig:occlusions} (right)) that surround the objects as inpaint masks
during training. Since we have GT background RGB along with estimated background depth within occlusion masks, we can directly use these masks along with the original image as GT for training. 

The intuition behind inpainting the occlusion mask is to pretend that the FG is larger than its actual size along its silhouette.
We find training on these masks helps the model learn to borrow from the regions with larger depth values. In other words, such training with occlusion masks makes inpainting depth-aware. 
Training with this type of mask only, however, is insufficient as the model has not learned to inpaint thin objects or perform regular inpainting. We address this by randomly adding traditional stroke-shape inpainting masks used in standard inpainting training following Deepfillv2 ~\cite{yu:2019:gated_convolution_inpainting}, which enables the model to learn to inpaint thin or small objects.  So our dataset consists of two types of inpainting masks: occlusion masks and random strokes. In this way, any single image dataset can be adapted to be used in training our inpainting model without requiring any annotations.
We show examples from our custom training dataset in the supplement. See Figure~\ref{fig:depth-aware-inpainting} for a sample inpainting result with more in the supplementary material.

We employ a patch-based discriminator \textbf{$D$} to discriminate between real and generated results, and apply an adversarial loss to the inpainting network, as in Deepfillv2 ~\cite{yu:2019:gated_convolution_inpainting}. So the objective loss for the inpainting network is a weighted sum of the reconstruction loss -- the $L_1$ distance between inpainted results and ground truth -- and the hinge adversarial loss. More details about network training and architectures are discussed in the supplementary material.

It is worth contrasting our inpainting approach with that of 3D-Photo inpainting~\cite{shih20203d}, which is also depth-aware and its dataset is annotation-free. One big difference is that our inpainting is global while 3D-Photo inpaints on local patches based on depth edges. Due to this, our inference only requires a single-pass and is relatively fast, while 3D photos requires multi-stage processing and iterative flood-fill like algorithms to generate the inpainting masks per patch, which is relatively time-consuming.

\subsection{Layered Rendering}
\label{sec:rendering}

Given the foreground and background images and disparity maps, we can now render each into a novel view and composite them together.  The soft layering stage produces a foreground layer comprised of the input image $I$, visibility map $A$, and disparity $D$.  We back-project the disparity map in the standard way to recover a 3D point per pixel and connect points that neighbor each other on the 2D pixel grid to construct a triangle mesh.  We then texture this mesh with $I$ and $A$ and render it into the novel viewpoint; $A$ is resampled, but not used for compositing during rendering at this stage.  The novel viewpoint is given by a rigid transformation $T$ from the canonical viewpoint, and the result of this rendering step is a new foreground RGB image $I_T$ and visibility $A_T$.  The output of the inpainting stage is background image $\tilde{I}$ and disparity $\tilde{D}$.  We similarly construct a triangle mesh from $\tilde{D}$, texture it with $\tilde{I}$, and project into the novel view to generate new background image $\tilde{I}_T$.  Finally, we composite foreground over background to obtain the final novel-view image $I_T^*$:
\vspace{-1mm}
\begin{equation}
  I_T^* = A_T I_T + (1 - A_T)  \tilde{I}_T.
  \vspace{-1mm}
\end{equation}
We use a TensorFlow differentiable renderer~\cite{genova2018unsupervised} to generate $I_T$ and $\tilde{I}_T$ to enable a unified framework.

\section{Experiments}
\label{sec:experiments}
\vspace{-2mm}

We quantitatively evaluate SLIDE on three multi-view
datasets: RealEstate10k~\cite{zhou2018stereo} (RE10K), Dual-Pixels~\cite{garg2019learning} 
and Mannequin-Challenge (MC)~\cite{li:2019:moving_people} that provide videos 
or multi-view images of a static scene. In addition, we perform user studies  
on the photographs from Unsplash~\cite{unsplash}. 

\vspace{1mm}
\noindent \textbf{Baselines and Metrics.}
We quantitatively compared with three recent state-of-the-art
techniques, for which code is publicly available:
SynSin~\cite{wiles2020synsin}, Single-image MPI~\cite{tucker2020single}~(SMPI)
and 3D-Photo~\cite{shih20203d}. SynSin and SMPI are end-to-end trained
networks that take a single image as input and generate novel-view
images. 3D-Photo, on the other hand, is a modular approach that
is not end-to-end trainable. Like SLIDE, 3D-Photo uses a disparity
network coupled with a specialized inpainting network to generate
novel-view images. For fair comparison, both 3D-Photo and SLIDE techniques
use MiDaSv2~\cite{lasinger:2019:depth_3d} disparities.
But, unlike SLIDE, 3D-Photo does not model
fine structures, like fur and hair, on foreground silhouettes.
We refer to our model that does not use alpha mattes as `SLIDE' 
and the one that does as `SLIDE with Matte'.

Following SMPI~\cite{tucker2020single}, we quantitatively measured the accuracy of the 
predicted target views with respect to the ground-truth images using 
three different metrics of LPIPS~\cite{zhang:2018:lpips}, PSNR and SSIM.
Since SLIDE and several baselines do not perform explicit out-painting
(in-filling the newly exposed border regions), we ignore a 20\%
border region when computing the metrics.

\begin{figure*}[t!]
\begin{minipage}{\textwidth}
    \centering
    \vspace{-4mm}
    \includegraphics[width=.95\textwidth]{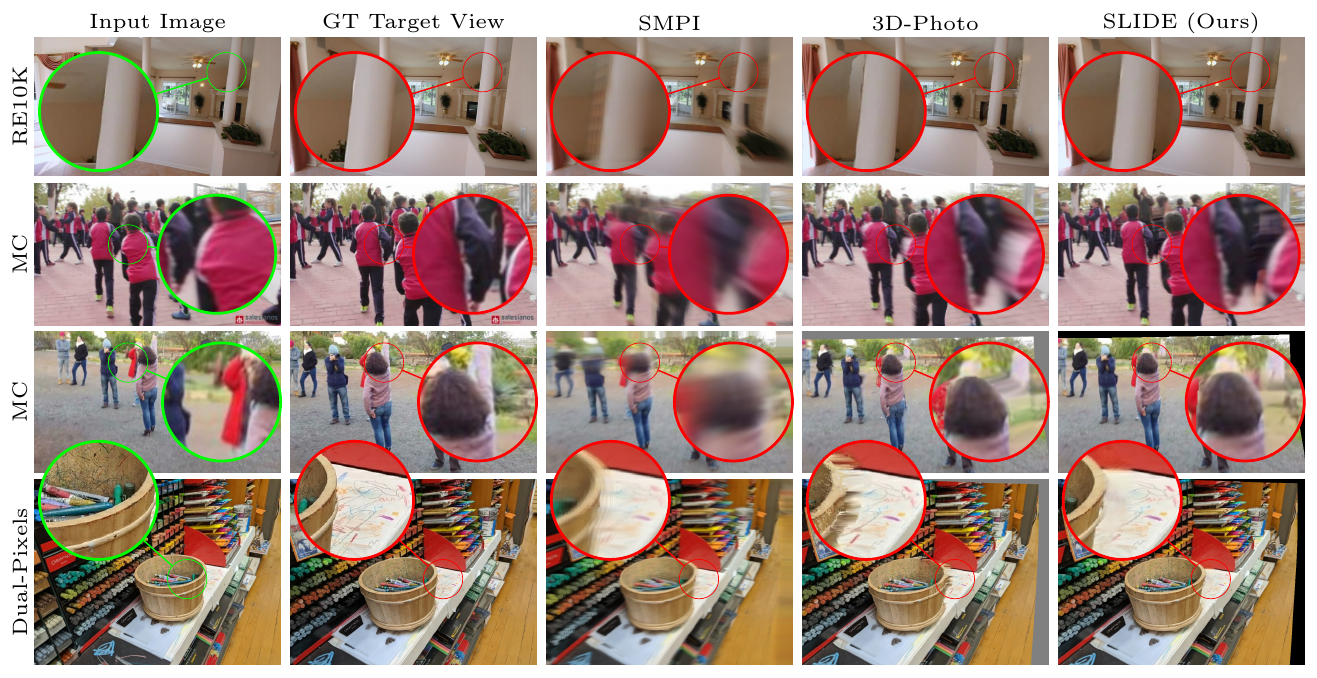}
    \caption{\textbf{Sample Visual Results on Benchmarks.} Novel view synthesis results of different techniques: Single-image MPI~\cite{tucker2020single}, 3D-Photo~\cite{shih20203d} and SLIDE (Ours) on sample images from RE10K~\cite{zhou2018stereo}, MC~\cite{li:2019:moving_people} and Dual-Pixels~\cite{garg2019learning} datasets.
    }
    \label{fig:visual_results_1}
    \vspace{.5mm}
\end{minipage}
\begin{minipage}{\textwidth}
    \centering
    \vspace{-2mm}
    \includegraphics[width=.95\textwidth]{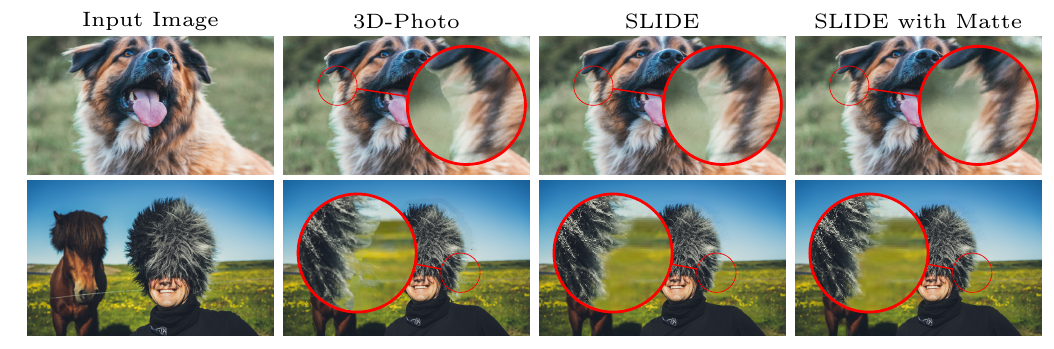}
    \caption{\textbf{Visual Results on in-the-wild Images.} View synthesis results on sample Unsplash dataset~\cite{unsplash}
    images that we use in user studies. SLIDE and SLIDE with Matte approaches can more faithfully represent thin hair-like
    structures compared to 3D-Photo~\cite{shih20203d}. See the supplementary video for better illustration of results.
    }
    \label{fig:visual_results_matting}
    \vspace{-4mm}
\end{minipage}    
\end{figure*}

\vspace{1mm}
\noindent \textbf{Results on RealEstate10K.}
RealEstate10K (RE10K)~\cite{zhou2018stereo} is a video clips dataset
with around 10K YouTube videos of static scenes. 
We use 1K randomly sampled video clips from the test set 
for evaluations. We follow~\cite{tucker2020single} and use structure-from-motion
and SLAM (Simultaneous Localization and Mapping) algorithms to recover camera intrinsics and extrinsics
along with the sparse point clouds. As in~\cite{tucker2020single}, we also 
compute the point-visibility from each frame. 
For the evaluation, we randomly sample a source view from each test clip and consider the following
$5^{th}$ ($t=5$) and $10^{th}$ ($t=10$) frames as target views.
We compute evaluation metrics with respect to these target views.
Table~\ref{table:re10k_results} shows that SLIDE performs better or on-par with current
state-of-the-art techniques with respect to all the evaluation metrics.
The improvement is especially considerable in the LPIPS perceptual metric, indicating that
SLIDE view synthesis preserves the overall scene content better than existing techniques.
Figure~\ref{fig:visual_results_1} shows sample visual results. 
SMPI~\cite{tucker2020single} generate blurrier novel views; and SLIDE usually preserves the structures better around the occlusion
boundaries where the scene elements on images move with respect to each other when camera moves.
On all the 3 benchmark datasets, we do not see further improvements by incorporating FG matting into SLIDE (SLIDE with Matte),
as these dataset images do not have predominant foreground objects with thin hair-like structures.

\vspace{1mm}
\noindent \textbf{Results on Dual-Pixels.}
Dual-Pixels~\cite{garg2019learning} is a multi-view dataset taken 
with a custom-made, hand-held capture rig consisting of 5 mobile phones. 
That is, each scene is captured simultaneously with 5 cameras that are separated by a moderate
baseline. 
We evaluate SLIDE and other baselines on the 684 publicly available test scenes.
For each scene, we consider one of the side views as input and consider the remaining 4 views
as target views.
Compared to RE10K, dual-pixels data consists of more challenging scenes captured in diverse settings. 
Table~\ref{table:dp_results} shows the quantitative results on the dual-pixels test dataset.
Results show that SLIDE outperforms all three baselines.
Figure~\ref{fig:visual_results_1} shows some qualitative results.

\vspace{1mm}
\noindent \textbf{Results on Mannequin Challenge (MC).}
MC is a video dataset collected and processed in a similar fashion as RE10K dataset.
It contains videos of people performing the ``mannequin challenge,'' in which subjects attempt to hold still as a camera moves through the scene. This (near) static setup enables the use of standard structure-from-motion pipelines,
like in RE10K, to obtain ground-truth camera poses and rough 3D point clouds of the scenes.
MC provides a good benchmark for view synthesis in scenes with people. We randomly sampled 190 5-frame sequences from the publicly available test set
for our evaluation purposes. We treat the first frame as input and the remaining four as target views.
Table~\ref{table:mc_results} shows the average metrics for the different single-image 3D photography techniques.
Again we observe that SLIDE outperforms the other techniques.
Figure~\ref{fig:visual_results_1} shows some qualitative results.

\begin{table}[t]
\centering
\scriptsize
\begin{tabular}{lcccccc}
\toprule
& \multicolumn{2}{c}{LPIPS~$\downarrow$} & \multicolumn{2}{c}{PSNR~$\uparrow$} & \multicolumn{2}{c}{SSIM~$\uparrow$} \\
\cmidrule(lr){2-3}
\cmidrule(lr){4-5}
\cmidrule(lr){6-7}
Method  & $t=5$ & $t=10$ & $t=5$ & $t=10$ &  $t=5$ & $t=10$  \\ 
\midrule
SynSin~\cite{wiles2020synsin} & 0.31 & 0.34 & 22.7 & 20.6 & 0.72 & 0.67 \\
SMPI~\cite{tucker2020single} & 0.14 &  0.19 & 26.7 & 24.1 & 0.86 & 0.80 \\
3D-Photo~\cite{shih20203d} & 0.09 & 0.12 & 26.9 & \textbf{23.7} & \textbf{0.87} & \textbf{0.80} \\
\midrule
SLIDE (Ours) & \textbf{0.06} & \textbf{0.10} & \textbf{27.1} & \textbf{23.7} & \textbf{0.87} & \textbf{0.80} \\
\bottomrule
\end{tabular}%
\vspace{1mm}
\caption{\textbf{Results on RE10K.} LPIPS~\cite{zhang:2018:lpips}, PSNR and SSIM scores of different
techniques computed w.r.t. target views at two time steps $t=5$ and $t=10$.}
\vspace{-3mm}
\label{table:re10k_results}
\end{table}

\begin{table}[t]
\centering
\scriptsize
\begin{tabular}{lccc}
\toprule
& LPIPS~$\downarrow$ & PSNR~$\uparrow$ & SSIM~$\uparrow$ \\
\midrule
SynSin~\cite{wiles2020synsin} & 0.75 & 12.9 & 0.31  \\
SMPI~\cite{tucker2020single} & 0.49 & 16.3 & 0.42 \\
3D-Photo~\cite{shih20203d} & 0.27 & 16.3 & 0.42 \\
\midrule
SLIDE (Ours) & \textbf{0.23} & \textbf{16.8} & \textbf{0.44} \\
\bottomrule
\end{tabular}%
\caption{\textbf{Results on Dual Pixels.} LPIPS~\cite{zhang:2018:lpips}, PSNR and SSIM scores of different
techniques computed w.r.t. four target views in the 684 test data scenes.}
\vspace{-3mm}
\label{table:dp_results}
\end{table}

\begin{table}[t]
\centering
\scriptsize
\begin{tabular}{lccc}
\toprule
& LPIPS~$\downarrow$ & PSNR~$\uparrow$ & SSIM~$\uparrow$ \\
\midrule
SynSin~\cite{wiles2020synsin} & 0.33 & 16.9 & 0.53  \\
SMPI~\cite{tucker2020single} & 0.31 & 17.8 & 0.59 \\
3D-Photo~\cite{shih20203d} & 0.22 & 18.8 & 0.60 \\
\midrule
SLIDE (Ours) & \textbf{0.18} & \textbf{20.0} & \textbf{0.66} \\
\bottomrule
\end{tabular}%
\caption{\textbf{Results on Mannequin Challenge dataset.} LPIPS~\cite{zhang:2018:lpips}, PSNR and SSIM scores of different
techniques computed w.r.t. four target views.}
\vspace{-2mm}
\label{table:mc_results}
\end{table}

\begin{figure}[t!]
\centering
\vspace{-2mm}
\includegraphics[width=\columnwidth]{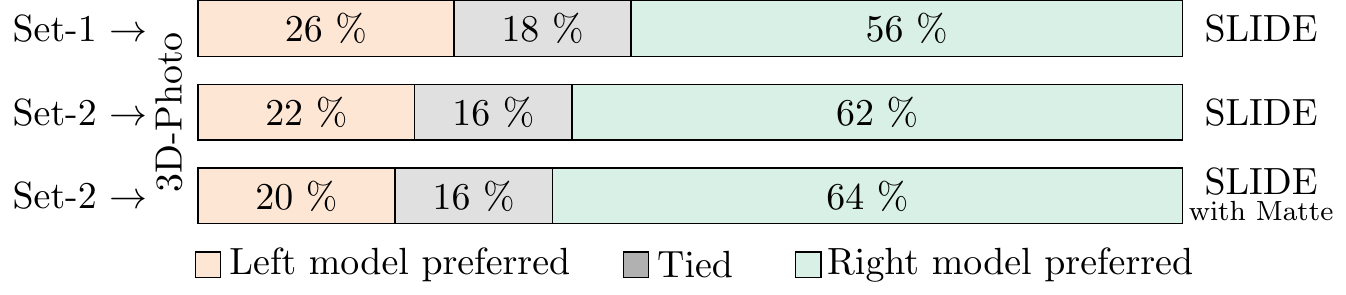}
\caption{\textbf{User Study.} Percentage of results users preferred between 3D-Photo and SLIDE results.
SLIDE results are consistently preferred on both sets of in-the-wild images.}
\vspace{-2mm}
\label{fig:user_study}
\end{figure}

\vspace{1mm}
\noindent \textbf{User Studies on in-the-wild images.}
In order to evaluate different techniques on ``in-the-wild'' photographs, we assembled two image sets
from Unsplash~\cite{unsplash}, a database of free-licensed, high-quality photographs taken by
professional photographers and hobbyists.  As we had no GT for these images, we performed
user studies to compare view-synthesis results of different techniques.
For the first set (Set-1), we collected 99 images with the elements we usually
see in photo albums: people, animals, vehicles and landscapes; sometimes
with multiple instances of these elements.
For the second set (Set-2), to demonstrate the use of matting,
we collected 50 images with close-ups of people and animals
with thin hair structures.
For each image in both sets, we created short videos with views synthesized from
circular camera paths 
(Please refer to the supplementary video to see sample generated videos).  
We then showed side-by-side results of our (SLIDE) method and a baseline method (in randomized pair order) to Amazon Mechanical Turkers and asked the user to choose the better looking video.
At least 15 users rated each video pair, and we took the majority vote
to compute the percentage of time users preferred one result over another.
User studies show SLIDE was preferred over SynSin~\cite{wiles2020synsin}
or SMPI~\cite{tucker2020single} over 99\% of time.
This result is not surprising, as these approaches usually produce blurrier results
than depth-based rendering approaches like SLIDE and 3D-Photo~\cite{shih20203d}.
Figure~\ref{fig:user_study} shows the result of comparing SLIDE and `SLIDE with Matte'
to 3D-Photos. On both image sets, users prefer SLIDE considerably more
often than 3D-Photo.
SLIDE is preferred still more on Set-2 images with thin hair-like structures, which are not well-handled by 3D-Photo, with additional gains when adding matting to the SLIDE pipeline (`SLIDE with Matte').
Figure~\ref{fig:visual_results_matting} shows some sample view synthesis results with
more in the supplementary material.

\vspace{1mm}
\noindent \textbf{Runtime Analysis.}
The SLIDE framework requires only a single forward pass
over all its components to generate the two layer representation, which can then be rendered in realtime on modern graphics chips to synthesize new views. All of the SLIDE components
can be realized using standard GPU layers in modern deep learning toolboxes resulting in
a unified and efficient system. We implemented most of the SLIDE components in Tensorflow except the 
depth estimation, foreground saliency and alpha matting networks, for which we used the original
PyTorch networks. The following are the runtimes of different SLIDE components when processing
a $672 \times 1008$ image on an Nvidia Titan P100 GPU: 
Depth estimation~\cite{lasinger:2019:depth_3d} (0.023s), Soft-layering (0.013s),
Depth-aware inpainting (0.037s). And optionally for matting: foreground segmentation~\cite{qin2020u2} (0.069s) and FBA-matting~\cite{forte2020f} (0.203s), for a total runtime of 0.07s or 0.35s (w/o or w/ matting, resp.).  By comparison, the closest competitor in quality, 3D-Photo, takes several seconds to process a single image.

\vspace{-2mm}
\section{Conclusion}
\vspace{-2mm}
SLIDE is a modular yet unified approach for 3D photography
that has several favorable properties: Soft layering that can model intricate 
appearance details; depth-aware inpainting; modular and unified system
with efficient runtime; and state-of-the-art results.
We observe artifacts when a component such as depth estimation
or alpha matting fails. 
We believe that the SLIDE 3D photographs will get better as these
components become even more mature and robust.

\vspace{2mm}
\small
\noindent \textbf{Acknowledgments.}
We thank Forrester Cole and Daniel Vlasic for their help with differentiable
rendering. We thank Noah Snavely and anonymous reviewers for their valuable
feedback.

{\small
\bibliographystyle{ieee_fullname}
\bibliography{refs}
}

\newpage

\part*{Supplementary Material}

\setcounter{table}{0}
\setcounter{figure}{0}
\setcounter{section}{0}
\renewcommand{\thetable}{T\arabic{table}}
\renewcommand{\thefigure}{S\arabic{figure}}
\renewcommand\thesection{A\arabic{section}}

In the supplementary material, we provide additional network and training details of our depth-aware inpainting network along with some preliminary results on end-to-end training of the SLIDE framework.
We recommend readers to view the video summary and additional results present in
the project page: \url{{https://varunjampani.github.io/slide}}.

\section{Details on Depth-aware RGBD Inpainting}
\label{sec:inpainting_supp}
\vspace{-2mm}
Here, we provide additional implementation details of our RGBD inpainting network, including network architectures, training hyperparameters, and also additional details of our training datasets.

\vspace{1mm}
\noindent \textbf{Network Architecture.}
We adopt the same coarse and refinement models as in \cite{DBLP:journals/corr/abs-1806-03589}. The main difference between our models and \cite{DBLP:journals/corr/abs-1806-03589} is the number of input and output channels for each model, as we do RGBD inpainting in constrast to RGB inpainting in~\cite{DBLP:journals/corr/abs-1806-03589}. We provide the network architecture details of 
the coarse inpainting model in Table~\ref{table:architecture}. We use the SN-PatchGAN~\cite{DBLP:journals/corr/abs-1806-03589} as the discriminator during training. 

\begin{figure}[th!]
\centering
\includegraphics[width=\columnwidth]{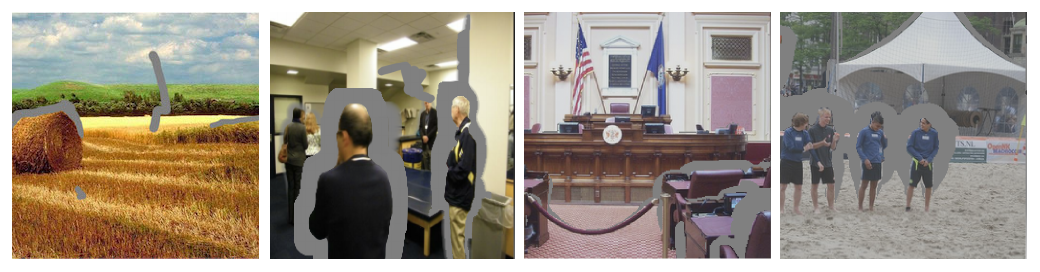}
\vspace{-1.5mm}
\caption{\textbf{Sample Inpainting Training Masks.} We use both occlusion masks as well as random strokes (shown as grey regions here) to train our RGBD inpainting network.}
\vspace{-2.5mm}
\label{fig:inpaint_masks}
\end{figure}

\vspace{1mm}
\noindent \textbf{Training Details.}
We train our inpainting network on images from Places2~\cite{DBLP:journals/corr/ZhouKLTO16}
dataset. As explained in the main paper, we use two types of masks for training our inpainting network: 
One is the occlusion masks that encourages the network to inpaint only from the background regions.
Another is the random strokes that are typically used in training a inpainting network.
Figure~\ref{fig:inpaint_masks} illustrates some sample inpaint masks used in training.
We implement our model using TensorFlow,
and trained for 4010K steps on image crops of resolution 256 x 256 with a batch size of 64. We randomly crop the images from 512 x 512 as the depth-aware inpainting is usually a local inpainting task. We also apply random flips and contrast augmentations during training. 

\begin{table}[h]
\centering
\scriptsize
\begin{tabular}{cccccc}
\toprule
 Layer & Filter size & Channel & Dilation & Stride & Relu \\
\midrule
 GatedConv & 5 & 48 & 1 & 2 & elu \\
GatedConv & 3 & 48 & 1 & 1 & elu \\
GatedConv & 3 & 96& 1 & 2 & elu \\
GatedConv & 3 & 96 & 1 & 1 & elu \\
GatedConv & 3 & 192 & 1 & 1 & elu \\
\midrule
GatedConv & 3 & 192 & 2 & 1 & elu \\
GatedConv & 3 & 192 & 4 & 1 & elu \\
GatedConv & 3 & 192 & 8 & 1 & elu \\
GatedConv & 3 & 192 & 16 & 1 & elu \\
\midrule
GatedConv & 3 & 192 & 1 & 1 & elu \\
GatedConv & 3 & 192 & 1 & 1 & elu \\
NearestUpsample x 2 &&&&&\\
GatedConv & 3 & 96 & 1 & 1 & elu \\
GatedConv & 3 & 96 & 1 & 1 & elu \\
NearestUpsample x 2 &&&&&\\
GatedConv & 3 & 48 & 1 & 1 & elu \\
GatedConv & 3 & 24 & 1 & 1 & elu \\    
GatedConv & 3 & 4 & 1 & 1 & elu \\   
\bottomrule
\end{tabular}%
\caption{\textbf{Inpainting Network Architecture.}}
\vspace{-3mm}
\label{table:architecture}
\end{table}

\begin{figure*}[t!]
\centering
\includegraphics[width=.95\textwidth]{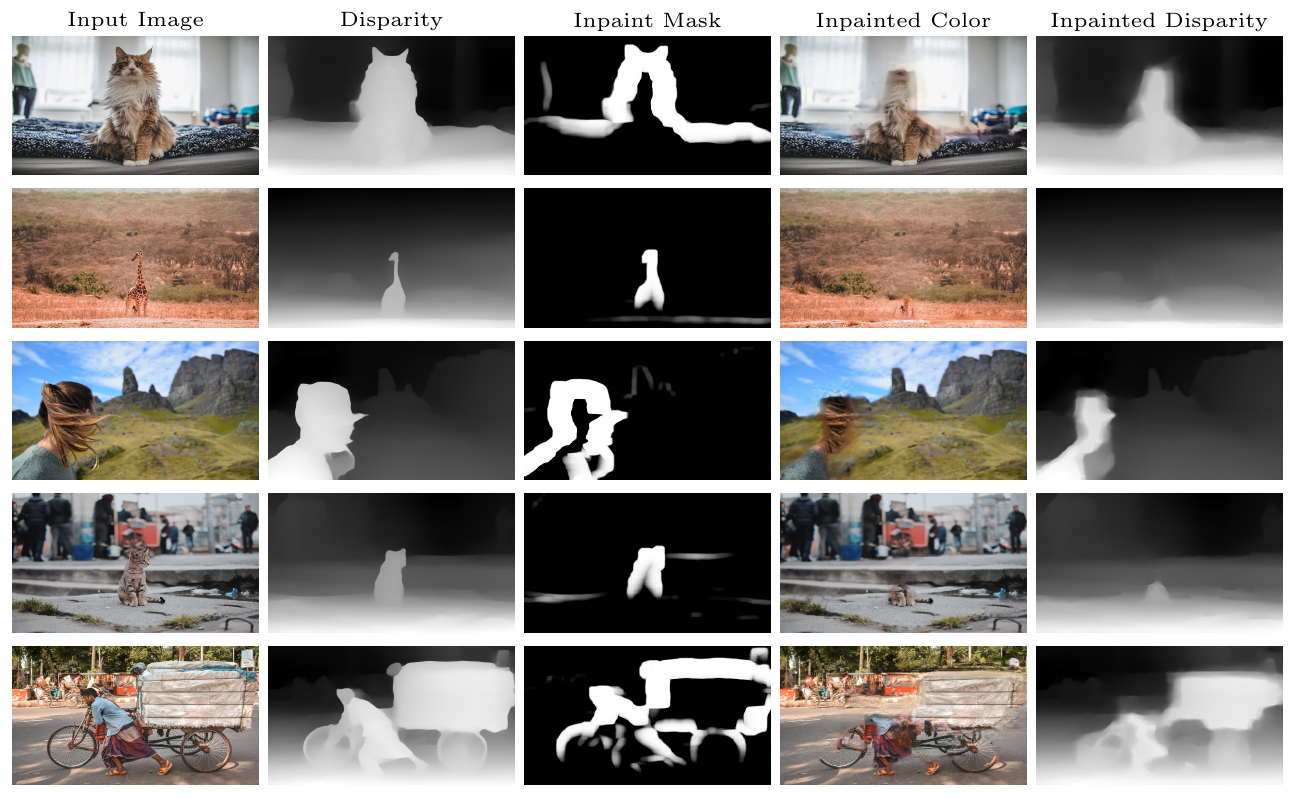}
\vspace{.5mm}
\caption{\textbf{Depth-aware Inpainting Results.} Sample visual results of inpainting demonstrates that our depth-aware RGBD inpainting borrows information predominantly from the background regions making it suitable for SLIDE 3D photography.}
\vspace{-1.5mm}
\label{fig:inpaint_visuals}
\end{figure*}

\vspace{1mm}
\noindent \textbf{Inpainting Visual Results.}
Figure~\ref{fig:inpaint_visuals} shows sample RGBD inpainting results where the inpaint mask
is the estimated disocclusions from the disparity map. Both RGB and disparity inpainting results
on diverse set of scenes show that the inpainted regions mostly borrow information
from the background regions. This demonstrate the effectiveness of our depth-aware inpainting
network.

\begin{table}[t]
\centering
\scriptsize
\begin{tabular}{lcccccc}
\toprule
& LPIPS~$\downarrow$ & PSNR~$\uparrow$ & SSIM~$\uparrow$ \\
\midrule
SynSin~\cite{wiles2020synsin}  & 0.34  & 20.6  & 0.67 \\
SMPI~\cite{tucker2020single}  &  0.19  & 24.1  & 0.80 \\
3D-Photo~\cite{shih20203d} & 0.12  & 23.7  & 0.80 \\
\midrule
\multicolumn{2}{c}{\textit{SLIDE variants}} & &  \\
SLIDE (Ours)  & 0.10  & 23.7  & 0.80 \\
with end-to-end fine-tuning  & 0.09 & 23.8  & 0.81 \\
\bottomrule
\end{tabular}%
\vspace{1mm}
\caption{\textbf{RE10K Results.} LPIPS [40], PSNR and SSIM scores of different
techniques computed w.r.t. target views ($t=10$).}
\vspace{-3mm}
\label{table:re10k_results_supp}
\end{table}

\section{End-to-end Training}
Though not reported in the paper, we actually have fine-tuned the model
with end-to-end training.
Table~\ref{table:re10k_results_supp} shows the (modest) gains obtained for one of the datasets (RE10K).  
Note that the multi-view datasets available to us do not have many 
large (dis)occlusions between views, which we suspect is needed 
to jointly supervise the depth and inpainting networks enough to 
achieve larger gains.
Note that large datasets are used to train depth ($\sim$1.9M images) and
inpaint ($\sim$2M images) networks in comparison to $\sim$10K scenes in RE10K.
Overall, SLIDE does enable end-to-end training with some gains, 
but further work on datasets and likely on loss functions is needed to maximize the benefit.
Even without end-to-end training, we believe
that SLIDE provides several insights 
with soft layering, (dis)occlusion reasoning, specialized RGBD inpainting
resulting in a fast, modular and unified framework.

\end{document}